\documentclass[10pt, conference]{IEEEtran}
\usepackage{cite}
\usepackage{amsmath,amssymb,amsfonts}
\usepackage{graphicx}
\usepackage{textcomp}
\usepackage{xcolor}
\usepackage{enumitem}
\setlist{leftmargin=6mm}
\def\BibTeX{{\rm B\kern-.05em{\sc i\kern-.025em b}\kern-.08em
    T\kern-.1667em\lower.7ex\hbox{E}\kern-.125emX}}
\usepackage{algpseudocode}          
\usepackage{algorithm}
\usepackage[flushleft]{threeparttable}

\usepackage{subfiles}
\newcommand\Mark[1]{\textsuperscript#1}
\makeatletter
\def\ps@IEEEtitlepagestyle{%
\def\@oddfoot{\mycopyrightnotice}%
\def\@evenfoot{}%
}
\def\mycopyrightnotice{%
{978-1-6654-3274-0/21/\$31.00~\copyright 2021 IEEE\hfill} 
\gdef\mycopyrightnotice{}
}
\begin{document}

\title{DNN-Opt: An RL Inspired Optimization for Analog Circuit Sizing using Deep Neural Networks
}

\author{\IEEEauthorblockN{Ahmet F. Budak\Mark{1}\IEEEauthorrefmark{1}, Prateek Bhansali\Mark{2}, Bo Liu\Mark{3}, Nan Sun\Mark{1}, David Z. Pan\Mark{1} and Chandramouli V. Kashyap\Mark{2}\IEEEauthorrefmark{2}}
	
\IEEEauthorblockA{\Mark{1}ECE Department, The University of Texas at Austin ~ \Mark{2}Intel Corp. ~  \Mark{3}James Watt School of Eng., University of Glasgow\\
\IEEEauthorrefmark{1} ahmetfarukbudak@utexas.edu,
\IEEEauthorrefmark{2}chandramouli.v.kashyap@intel.com
}
}

\maketitle

\begin{abstract}
Analog circuit sizing takes a significant amount of manual effort in a typical design cycle. With rapidly developing technology and tight schedules, bringing automated solutions for sizing has attracted great attention. This paper presents DNN-Opt, a Reinforcement Learning (RL) inspired Deep Neural Network (DNN) based black-box optimization framework for analog circuit sizing. The key contributions of this paper are a novel sample-efficient two-stage deep learning optimization framework leveraging RL actor-critic algorithms, and a recipe to extend it on large industrial circuits using critical device identification. Our method shows 5--30x sample efficiency compared to other black-box optimization methods both on small building blocks and on large industrial circuits with better performance metrics. To the best of our knowledge, this is the first application of DNN-based circuit sizing on industrial scale circuits.
\end{abstract}

\vspace{-1mm}
\section{Introduction}
Analog Integrated Circuit (IC) design is a complex process involving multiple steps. Billions of nanoscale transistor devices are fabricated on a silicon die and connected via intricate metal layers during those steps. The final product is an IC, which powers much of our life today. An essential aspect of IC design is analog design, which continues to suffer from long design cycles and high design complexity due to lack of automation in analog Electronic Design Automation (EDA) tools compared to digital flows. In particular, ``circuit sizing'' tends to consume a significant portion of analog designers' time. In order to tackle this labor-intensive nature and reduce time-to-market requirements, analog circuit sizing automation has attracted high interest in recent years. 

Prior work on analog circuit sizing automation can be divided into two categories: knowledge-based and optimization-based methods. In the knowledge-based approach, design experts transcribe their domain knowledge into algorithms and equations \cite{10.1023/A:1015098112015},\cite{JANGKRAJARNG2003237}. However, such methods create dependency on expert human-designers, circuit topology, and technology nodes. Thus, these methods are highly time-consuming and not scalable. 

Optimization-based methods are further categorized into two classes: equation-based and simulation-based methods. Equation-based methods try to express circuit performance via posynomial equations or regression models using simulation data. Then the equation-based optimization methods such as Geometric Programming\cite{1196196}, \cite{BoydOpAmpGP} or Semidefinite Programming (SDP) relaxations \cite{6881491} are applied to convex or non-convex formulated problems to find an optimal solution. Although those methods are generally fast, developing accurate expressions for circuit performances is not easy and deviates largely from the actual values. On the other hand, simulation-based methods employ black-box or learning-based optimization techniques to explore design space. These methods make guided exploration in the search space and target a global minimum using the real evaluations from circuit simulators. 

Traditionally, there have existed various model-free optimization methods such as particle swarm optimization (PSO)\cite{Vural2012AnalogCS} and advanced differential evolution \cite{ Liu:2013:ADA:2526263}. Although these methods have good convergence behavior, they are known to be sample-inefficient (i.e., SPICE simulation intensive). Recently surrogate model-based and learning-based methods are becoming increasingly popular due to their efficiency in exploring solution space. In surrogate model-based methods, Gaussian Process Regression (GPR)\cite{10.5555/1162254} is generally used for design space modeling, and the next design point is determined through model predictions. For example, GASPAD method is introduced into Radio Frequency (RF) IC synthesis where GPR predictions guide evolutionary search \cite{GASPAD}. WEIBO method proposed a GPR based Bayesian Optimization \cite{NIPS2012_05311655} algorithm where a blended version of weighted Expected Improvement (wEI) and the probability of feasibility is selected as acquisition function to handle constrained nature of analog sizing \cite{Lyu:2018:MBO:3195970.3196078}. The main drawback of Bayesian Optimization methods is scalability as GP modeling has cubic complexity in the number of samples, $\mathcal{O}(N^3)$.  

Recently, reinforcement learning algorithms are applied in the area as learning-based methods. GCN-RL\cite{Wang2020GCNRLCD} leverages Graph Neural Networks (GNN) and proposes a transferable framework. Despite reporting superior results over various methods and human-designer, a) it requires thousands of simulations for convergence (without transfer learning) and b) it suffers from engineering effort to determine observation vector, architecture selection, and reward engineering. AutoCkt \cite{Settaluri2020AutoCktDR} is a sparse sub-sampling RL technique optimizing the circuit parameters by taking discrete actions in the solution space. AutoCkt shows more efficiency over random RL agents and Differential Evolution. Still, it requires to be trained with thousands of SPICE simulations before deployment, which is costly. 

In this paper we introduce DNN-Opt, a two-stage deep learning black-box optimization scheme, where we merge the strengths of Reinforcement Learning (RL), Bayesian Optimization (BO), and population-based techniques in a novel way. The key features of the DNN-Opt framework are below.
\begin{itemize}
	\item We tailored a two-stage Deep Neural Network (DNN) architecture for black-box optimization tasks inspired by the actor-critic algorithms developed in the RL community.
	\item To leverage convergence behavior of population-based methods, DNN-Opt adopts a population-based search space control mechanism.
	
	\item We introduce a recipe for extending our work for large industrial designs using sensitivity analysis. In collaboration with a design house, we demonstrate that our work can also efficiently size large circuits with tens of thousands of devices in addition to small building blocks. 
\end{itemize}

The rest of the paper is organized as follows. We formulate analog circuit sizing problem in Section II and introduce DNN-Opt with its RL core and other details.  In Section III, the performance of DNN-Opt is demonstrated on small building blocks and large industrial circuits. We also provide performance comparisons of DNN-Opt with other optimization methods. The conclusions are provided in Section IV.


\section{DNN-Opt Framework}
\subsection{Analog Circuit Sizing: Problem Formulation}
We formulate analog circuit sizing task as a constrained optimization problem succinctly as below.
\begin{equation}
	\label{eq:prob_formulation}
	\begin{aligned} \operatorname{minimize}\text{ } & f_{0}(\mathbf{x}) \\ \text { subject to } & f_{i}(\mathbf{x}) \leq 0 \quad \text { for } i=1, \ldots, m 
	\end{aligned}
\end{equation}

where, $\mathbf{x}\in\mathbb{D}^{d}$ is the parameter vector and $d$ is the number of design variables of sizing task. Thus,  $\mathbb{D}^{d}$ is the design space. $f_0(\mathbf{x})$ is the objective performance metric we aim to minimize. Without loss of generality, we denote $i^\text{th}$ constraint by $f_i(\mathbf{x})$.
\subsection{DNN-Opt Core: RL Inspired Two-Stage DNN Architecture}
\begin{figure}
	\centering
	\includegraphics[scale=0.37]{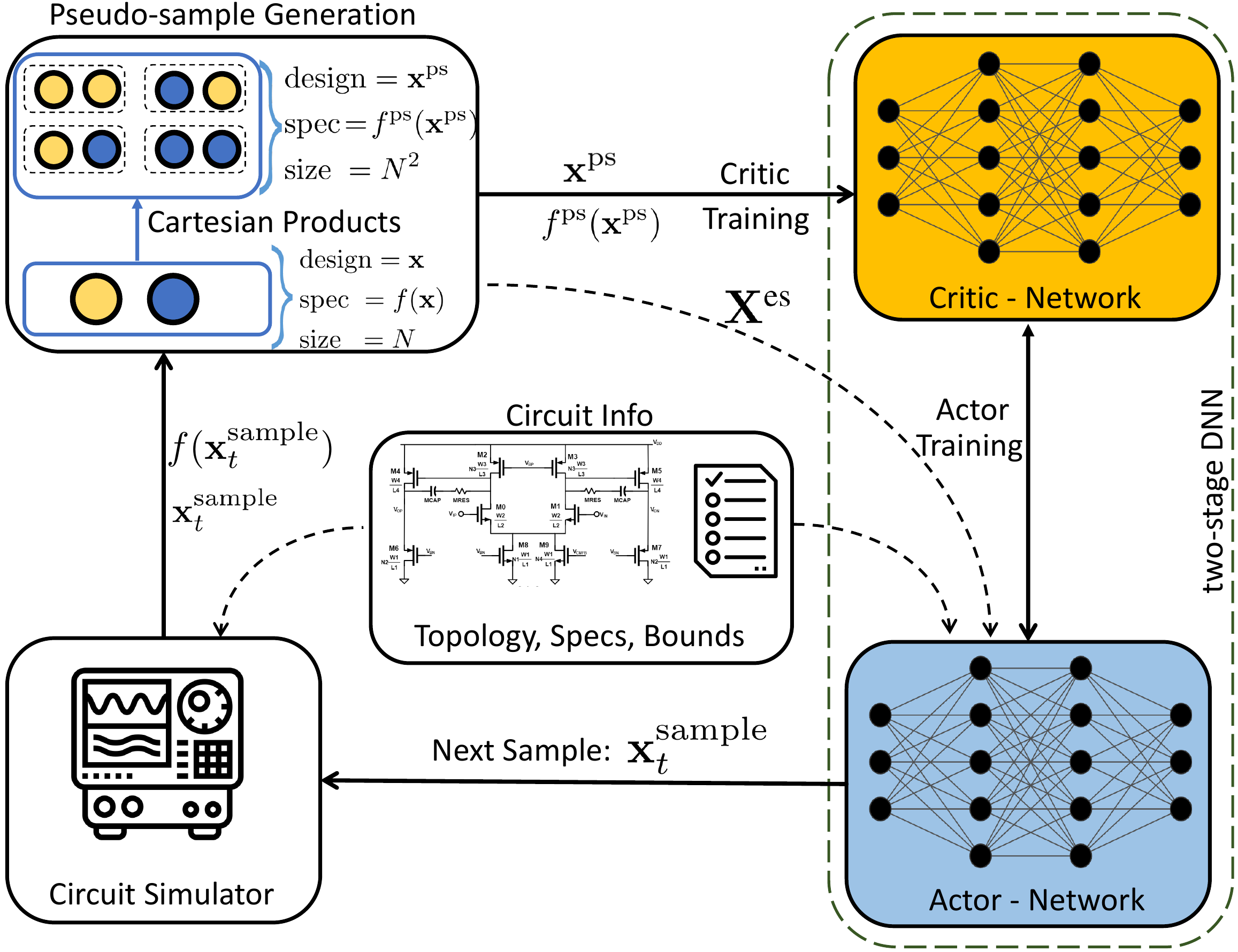}
	\vspace*{-3mm}
	\caption{DNN-Opt Framework}
	\label{fig:DNN_Opt_Core_final}
	\vspace*{-4mm}
\end{figure}

The overall framework of DNN-Opt is shown in Figure \ref{fig:DNN_Opt_Core_final}. DNN-Opt comprises a two-stage deep neural network architecture that interacts with a circuit simulator during the optimization process. The flow starts from generated samples in the design space; then, a critic-network is used to predict any new design point's performance. This prediction is used by the actor network to propose new candidates for simulation. This search scheme efficiently mimics BO behavior in space exploration. Besides, the sample generation is further optimized by adopting a population control scheme.

The two-stage network architecture of our work borrows its structure from  Deep Deterministic Policy Gradient (DDPG) algorithm \cite{journals/corr/LillicrapHPHETS15}, which is an RL actor-critic algorithm \cite{Konda00actor-criticalgorithms} developed for continuous action spaces. However, actor-critic algorithms are not directly applicable to analog circuit sizing since it is not a Markov Decision Processes (MDP) \cite{10.5555/3312046}, which is a \textit{necessary condition} for any RL problem. Therefore we adapt DDPG algorithm with significant modifications tailored for analog circuit sizing.

In the context of analog circuit sizing, we will keep some of the RL notation but replace many for simplicity and clarity.\\
\textbf{Design}: A design is a set of circuit parameters which we denote by $\mathbf{x}$ and it is a vector of size $d$ where each element corresponds to a particular design variable. The optimization goal is to find optimal $\mathbf{x}_{\text{opt}}$ which satisfies Eq. \ref{eq:prob_formulation}.
\\
\textbf{Population}: A population is set of multiple designs.
\\
\textbf{Design Population Matrix}:  We define a design population matrix as $\mathbf{X} \in \mathbb{R}^{N{\times}d}$, where $N$ is the population size. The parameters of $i^\mathrm{th}$ design is a row in the design population matrix $\mathbf{X}$, which is denoted as $\mathbf{x}_i$.
\\
\textbf{State Space}: Our work maps optimization parameters (circuit design variables) to state representation in RL notation. A state of $k^\text{th}$ design is transformed as $\textbf{s}_{k} = \mathbf{x}_k$.\\
\textbf{Action Space}: Each action $\textbf{a}_{k}$ in our new architecture corresponds to \textit{change} in optimization parameters vector, $\mathbf{x}_k$, which can be denoted as $\textbf{a}_{k} = \Delta\mathbf{x}_k$. An intuitive explanation of this choice is that an ideal action for an optimization task should propose change in each design variable to have a better design.
\\
\textbf{Critic-Network}: Originally, a critic-network parameterized by $\theta^Q$ approximates the return value of an MDP $\mathrm{Return} = Q(s_t,a_t | \theta^Q)$. We modify its role and use this network as a proxy in lieu of expensive SPICE simulator. Our modified critic-network provides a vector-to-vector mapping by taking an $(\mathbf{x}, \Delta \mathbf{x}) \in \mathbb{D}^{2d}$ as input and providing performance predictions $Q(\mathbf{x},\Delta \mathbf{x} | \theta^Q) \in \mathbb{R}^{m+1}$ at output, one-dimension is for objective specification and m for constraint specifications.\\
\textbf{Actor-Network}: An actor-network parameterized by $\theta^\mu$ would take a state as its input and determine an action to take $\mathbf{a}_k = \mu(\mathbf{s}_k | \theta^\mu)$. In the context of analog circuit sizing, actor-network provides change in design parameter vector for design $k$ as: $\Delta \mathbf{x}_k = \mathbf{a}_k = \mu(\mathbf{x}_k | \theta^\mu)$.\\
\textbf{Critic-Network Training}: We utilize critic-network for modeling design variable to circuit performance relationship. For effective training, we use data augmentation techniques to generate $N^2$ \textit{pseudo-samples} $\mathrm{(ps)}$ using original $N$ samples. In order to generate pseudo-samples, we use two-samples $\mathbf{x}_i \text{ and }\mathbf{x}_j$ and corresponding spec vectors $f(\mathbf{x}_i) \text{ and }f(\mathbf{x}_j)$, as follows:
\begin{equation}
	\vspace{-2mm}
	\label{eq:pseudo_smp}
	\begin{aligned}
		&\mathbf{x}^\mathrm{ps}_{ij} = \left[\mathbf{x}_{i}, \Delta \mathbf{x}_{ij}\right] = \left[\mathbf{x}_i, \mathbf{x}_j - \mathbf{x}_i\right] \\ &f^{\mathrm{ps}}(\mathbf{x}^\mathrm{ps}_{ij}) = f(\mathbf{x}_{j})
	\end{aligned}
\end{equation}

This leads to change in the input dimensionality of critic-network from $d$ to $2d$ since we now have to use $(\mathbf{x}, \Delta \mathbf{x})$ instead of $\mathbf{x}\text{ or } \mathbf{(x + } \Delta \mathbf{x})$.
Our experiments conducted on Bayesmark \cite{BayesMark} benchmark problems showed that using $\mathrm{2}d$ inputs and training with pseudo-samples boosted critic-network's accuracy significantly over a network trained with $d$ inputs and original samples.

For a batch-size of $N_b$ pseudo-samples, the following Mean Squared Error (MSE) loss function is used to train the critic network.
\begin{equation}
	\label{eq:train_crit}
	\resizebox{0.91\columnwidth}{!}{%
		$L\left(\theta^{Q}\right)=\frac{1}{N_b (m+1)} \sum^{N_b}_{k=1}\sum^{{m+1}}_{l=1} \left( Q(\mathbf{x}_k, \Delta \mathbf{x}_k)^l - f(\mathbf{x}_k + \Delta \mathbf{x}_k)^l\right)^2
		$}
\end{equation}
where $Q(\mathbf{x}_k, \Delta \mathbf{x}_k)^l$ is the critic-network's approximation for $\mathrm{k}^{th}$ pseudo-sample's $l^{th}$ performance and $f(\mathbf{x}_k + \Delta \mathbf{x}_k)^l$ is the SPICE simulated value for the same design-performance pair. To clarify, we have SPICE simulation values for pseudo-samples because the way they are constructed. \\
\textbf{Actor-Network Training}:
Training of actor-network is done after critic-network is trained and its hyperparameters are fixed. The training of actor-network corresponds to search in design space for \textit{better} designs. We come up with a Figure of Merit (FoM) function, $g(\cdot)$, based on performance-vector to objectively quantify how better a design is with respect to others.
\begin{equation}
	\label{eq:scalarizationfunc}
	g\left[f(\mathbf{x})\right] = w_0\times f_0(\mathbf{x}) + \sum_{i=1}^{{m}} \mathrm{min}\left(1, \mathrm{max}(0, w_i \times f_i(\mathbf{x}))\right)
\end{equation}
where $w_i$ is the  weighting factor. Note, a $\mathrm{max(\cdot)}$ clipping used for equating designs after constraint are met and $\mathrm{min(\cdot)}$ clipping is used for practical purposes to prevent single constraint violation to dominate $g(\cdot)$ value. We train actor-network parameters by using $g(\cdot)$ function and replacing SPICE simulation values $f(\cdot)$ by the critic-network predictions $Q(\mathbf{x},\Delta\mathbf{x})$. We will further use a population of ``elite" solutions (es) of size $N_{\text{es}}$ to restrict search space for actor network. Population of elite solutions is a subset of total population determined based on the FoM ranking.

For a batch-size of $N_b$ samples the following loss-function is used to train actor network.
\begin{equation}
	\label{eq:train_actor}
	L\left(\theta^{\mu}\right) = \frac{1}{N_b} \sum^{N_b}_{k=1} \left(g\left[ Q(\mathbf{x}_k, \mu(\mathbf{x}_k \mid \theta^\mu)) \right] + \Vert\lambda*
	\mathrm{viol}_k\Vert_2\right)
\end{equation}
\noindent where $\mu(\mathbf{x}_k \mid \theta^\mu)$ is proposed parameter change vector $\Delta \mathbf{x}_k$ by the actor network. $\left(\lambda * \mathrm{viol}_k\right)$ is an element-wise vector multiplication where $\lambda$ is weighting coefficient chosen to be very large to prevent any boundary violation and keep the search in the restricted search region. The total boundary violation $\mathrm{viol}_k$ for action $k$ is defined as follows:
\begin{equation}
	\resizebox{0.91\columnwidth}{!}{%
		$\mathrm{viol}_k = \mathrm{max}(0,lb_{\mathrm{rest}} - (\mathbf{x}_k + \Delta \mathbf{x}_k)) + \mathrm{max}(0, (\mathbf{x}_k + \Delta \mathbf{x}_k) - ub_{\mathrm{rest}})  $
	}
\end{equation}
where $lb_{\mathrm{rest}}$ and $ub_{\mathrm{rest}}$ are the restriction boundary vectors for design variables determined by the population of elite solutions given by:
$$
\begin{aligned}
	lb^i_{\mathrm{rest}} =& \mathrm{min}(\mathbf{x}^i) \;\; \forall  i =1, \ldots,d \\
	ub^i_{\mathrm{rest}} =& \mathrm{max}(\mathbf{x}^i) \;\; \forall i =1, \ldots,d
\end{aligned}
$$
where, $\mathbf{x}^i$ is the column vector of size $N_{es}$ consisting of $i^{th}$ parameter of all designs in the elite population.

The hyperparameters (number of layers, number of nodes, learning rate, etc.) of the architecture for the actor and critic networks were found based on empirical studies.

\subsection{Sensitivity Analysis}
We use sensitivity analysis to prune design search space for efficiently finding an optimized solution. A blind search space exploration may lead to wasted circuit simulations during optimization. For example, in a classical seven transistor Operational Amplifier (OpAmp) \cite{BoydOpAmpGP} power dissipation does not depend on the differential pair devices once they are in saturation. Thus, if we want to size a circuit for reducing power, we should not make device properties of the differential pair devices as variables. To use sensitivity analysis in practice for any generic circuit, we first traverse the circuit hierarchy and collect all unique device design variables, $d$. Then, we perform sensitivity analysis by perturbing each of the design variables around its nominal value and observing its impact on objective and constraints, $f_i$. More formally, we compute sensitivity $\mathcal{S}_{ij}$ as 

\begin{equation}
	\label{eq:sens_equation}
	\mathcal{S}_{ij} = \frac{\delta f_i}{\delta d_j}, \forall i=0,\ldots,m ; j = 1, \ldots, d.
\end{equation}

We only need to consider design variables for which $\mathcal{S}_{ij} > thresh$, where $thresh$ is a user-defined number. Empirically, this analysis prunes design search space effectively, allowing us to work on large scale circuits.

We are now ready to present the overall framework of DNN-Opt in the next subsection.

\subsection{DNN-Opt: Overall Framework}
The overall framework for DNN-Opt is provided in Algorithm \ref{alg:dnnopt}. As a prerequisite, we apply sensitivity analysis for a large design and reduce number of design variables to a workable range. We then randomly sample $N_{\mathrm{init}}$ points from the design search space to build initial population. For optimization iteration $t$, first step is to initialize actor-critic parameters followed by pseudo-sample generation. Next actor-network and critic-network are trained. After this, an elite-population is constructed based on FoM of total-population (this elite-population will be updated with optimization iterations). The next query point is generated from elite-population, $\mathbf{X}^{\text{es}}$, using pre-trained actor-critic as follows. We use every design, $\mathbf{x}_{i}^{\text{es}}$, in the pool of elite-population as input to actor-network. The output of actor-network, $\Delta \mathbf{x}_i^\mathrm{es} = \mu(\mathbf{x}_{i}^{\mathrm{es}})$, is proposed change for design parameters in search of an optimal solution. With the imposed exploration noise $(\mathcal{N})$, a candidate design point is naturally formed as: $\mathbf{x}^\mathrm{ca}_i = \mathbf{x}^\text{es}_i + \mu(\mathbf{x}^{es}_i) + \mathcal{N}$. At this step, we have exactly the same number of proposed candidates, $\mathbf{X}^\text{ca} = [\mathbf{x}^\mathrm{ca}_i, \dots, \mathbf{x}^\mathrm{ca}_{N_\text{es}}]$, as the size of elite-population. Once  the population pairs, $\mathbf{X}^\text{es}$ and $\mathbf{X}^\text{ca}$, are formed the next sample point for iteration $t$ is selected using Eq. \ref{eq:find_query}.
\begin{equation}
	\label{eq:find_query}
	\resizebox{0.91\columnwidth}{!}{%
		$\mathbf{x}_{t}^\mathrm{sample} = \big[\mathbf{x}^\text{ca}_k \text{ for } k=arg\,min_i\left(g[Q(\mathbf{x}^\text{es}_i, \mathbf{x}^\text{ca}_i - \mathbf{x}^\text{es}_i)] \right)\big]
		$}
\end{equation}

\begin{algorithm}[h]
	\caption{DNN-Opt Algorithm}
	\label{alg:dnnopt}
	\begin{algorithmic}[1]
		\Require Dimensionality reduction with sensitivity analysis \textbf{if} design is \textit{large}
		\Require An initial sample set $\mathbf{X}^\mathrm{init}$ of $N_\mathrm{init}$ designs and their evaluations $f(\mathbf{X}^\mathrm{init})$
		\State Define total population $\mathbf{X}^\mathrm{tot} = \mathbf{X}^\mathrm{init}$
		\For {$t = 1, 2, \dots,t_{max}$}
		\State Initialize actor \& critic network parameters $\theta^\mu$ and $\theta^Q$
		\State Generate pseudo-samples using existing design $\mathbf{X}^\mathrm{tot}\rightarrow$ Eqn. \ref{eq:pseudo_smp}
		\State Train critic-network $\rightarrow$ Eqn. \ref{eq:train_crit}
		\State Train actor-network $\rightarrow$ Eqn. \ref{eq:train_actor}
		\State Calculate FoM for each design by $\text{FoM} = g[f(\mathbf{X}^\text{tot})]$
		\State Choose $N_\text{es}$ designs with smallest FoM to form population of elite solutions $\mathbf{X}^\mathrm{es}$.
		\State Find query point (next sample) $\mathbf{x}_t^\mathrm{sample}$ using actor-model $\rightarrow$ Eqn. \ref{eq:find_query}
		\State Simulate the query point and obtain specs $f(\mathbf{x}_t^\mathrm{sample})$ via SPICE sims
		\If {return cond(e.g. specs are met)} 
		\State break
		\EndIf
		\State $\mathbf{X}^\mathrm{tot}\text{.append}(\mathbf{x}_t^\mathrm{sample})$
		\State Go back to line 3
		\EndFor
		\State \Return The design with highest FoM
	\end{algorithmic}
	\vspace*{0mm}
\end{algorithm}
\section{Experimental Results \label{sec:experiments}}
\vspace{-1.5mm}
To demonstrate the reliability and efficiency of the DNN-Opt, we apply it to two sets of experiments using six circuit examples. The first experiment set is on small building blocks where every transistor is parameterized and sized, and the second experiment set includes larger industrial circuits with thousands of nodes and devices. 
\vspace{-2mm}
\subsection{Experiments with Small Building Blocks}
\vspace{-1mm}
\label{sec:small_building_blocks}
We tested DNN-Opt on two small building blocks: a folded cascode amplifier and a strong-arm latch comparator. We included the majority of the circuit performances in the constraint list to mimic real-world design experience. Both designs are implemented in 180nm CMOS technology.

We compare our algorithm with three other well-known methods: a) A Differential Evolution (DE) method, which is a conventional population-based model-free algorithm, b) Bayesian Optimization with weighted Expected Improvement (BO-wEI)\cite{Lyu:2018:MBO:3195970.3196078}, which is a modified version of Bayesian Optimization for constrained problems, and c) GASPAD method\cite{GASPAD}, a surrogate model (GP) assisted evolutionary framework. To account for the randomized techniques involved in all these methods, we repeat experiments ten times to report each method's findings. We determine the simulation budgets for our experiments by considering the convergence nature of the methods. DE has a simulation budget of 10000, and BO-wEI, GASPAD, and DNN-Opt are limited by 500 simulations. All the experiments are run on a workstation with Intel Xeon CPU and 128GB RAM, and a commercial SPICE simulator.
We used several metrics to compare the algorithms. We provide statistics of the methods for each example, and we denote the number of times a feasible solution is found by \textit{success rate}. We also share the evolution of FoM value calculated based on Eq. \ref{eq:scalarizationfunc} to demonstrate each algorithm's convergence during runtime. The constraint expressions given in Eq. \ref{specifications_folded} and \ref{specifications:SA} can be trivially readjusted to fit into the form of Eq. \ref{eq:prob_formulation}.

\begin{figure}[t]
	\vspace*{0mm}
	\centering
	\vspace*{0mm}
	\includegraphics[scale=0.45]{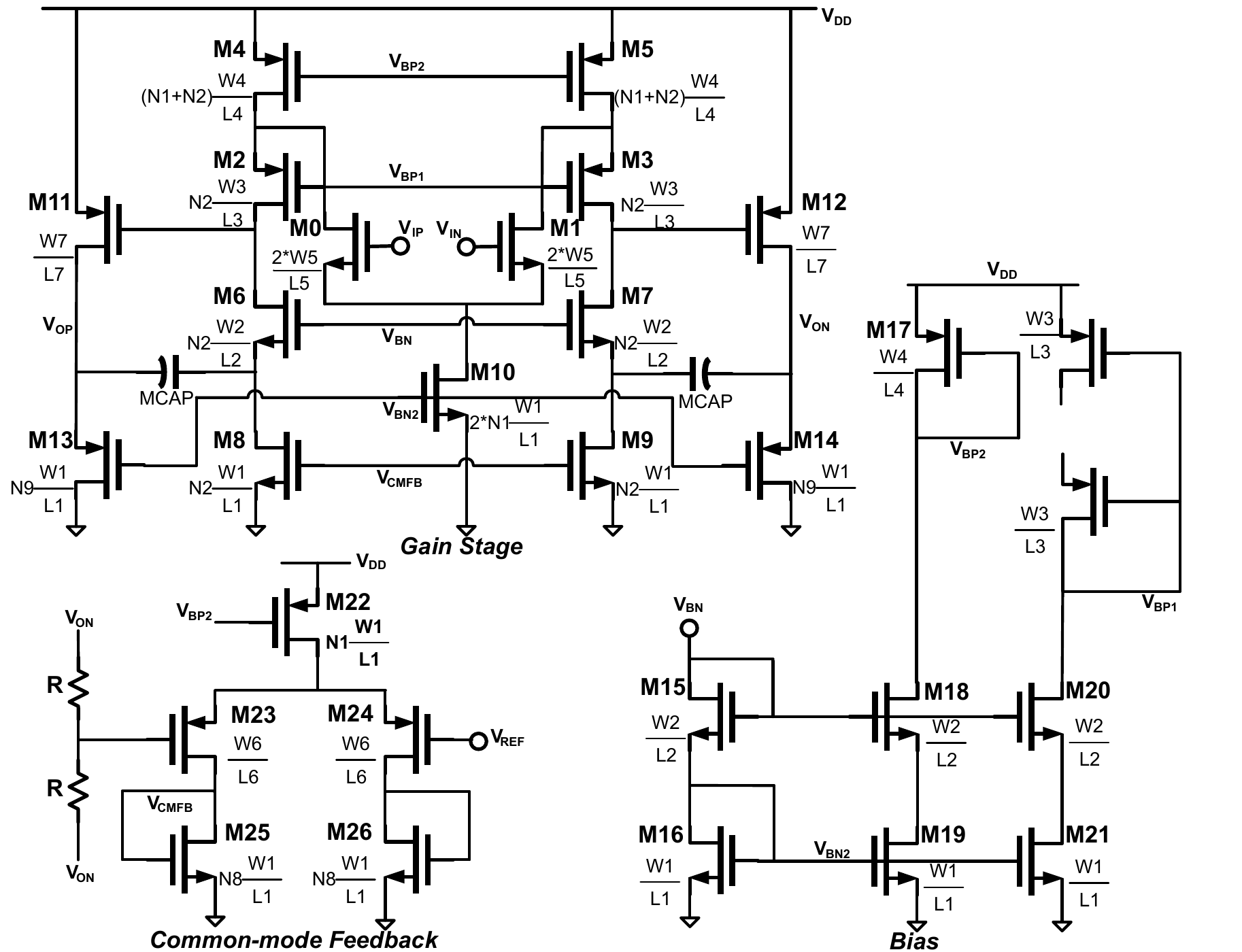}
	\vspace*{-2mm}
	\caption{Schematic of the folded-cascode OTA}
	\label{fig:foldedschematic}
	\vspace*{-4mm}
\end{figure}

\textbf{Folded Cascode OTA}:
The first test case is a two-stage folded-cascode Operational Transconductance Amplifier (OTA) (Figure \ref{fig:foldedschematic}).It has 20 design variables, and the designer provided search ranges are as shown in Table I.

\vspace*{-3mm}
\begin{table}[!h]
	\centering
	\caption{Design parameters and ranges for the folded-cascode OTA}
	\label{table:foldedspace}
	\vspace*{-4mm}
	\begin{center}
		\resizebox{0.9\columnwidth}{!}{%
			\begin{tabular}{l |l |l| l}
				\hline
				Parameter Name & Unit & LB & UB \\
				\hline
				L1-L2-L3-L4-L5-L6-L7& $\mu\text{m}$& 0.18 & 2\\
				\hline
				W1-W2-W3-W4-W5-W6-W7 \;\;\;& $ \mu\text{m}$& 0.24 & 150\\
				\hline
				N1-N2-N8-N9& integer & 1 & 20\\
				\hline
				MCAP & $f$$\text{F}$ & 100 & 2000\\
				\hline
				Cf & $f$ $\text{F}$ & 100 & 10000\\
				
				\hline
			\end{tabular}%
		}
	\end{center}
	\vspace*{-3mm}
	\begin{center}
		\begin{tablenotes}
			\small
			\item W:device width; L:device length; UB:upper bound; LB:lower bound
		\end{tablenotes}
	\end{center}
	\vspace*{-3mm}
\end{table}

The sizing problem is defined as follows:
\vspace*{0mm}
\small
\begin{equation}\label{specifications_folded}
	\resizebox{0.92\columnwidth}{!}{%
		$
		\begin{array}{l}
			{\text { minimize } \text{Power}} \\
			{\text { s.t. } \text{\enspace DC Gain} >60 \mathrm{\enspace dB} \qquad {\text{Settling Time}<\mathrm{30} \mathrm{\enspace ns}}} \\
			{\qquad \begin{array}{l l}
					
					{\text{CMRR}>80 \mathrm{ dB}} & {\text{Saturation Margin}>\mathrm{50 \enspace mV}} \\
					{\text{PSRR}>80 \mathrm{\enspace dB}} & {\text{Unity Gain Freq.}>30 \mathrm{\enspace MHz}}\\
					{\text{Out. Swing}>2.4 \mathrm{\enspace V}} & {\text{Out. Noise}<\mathrm{30} \mathrm{\enspace mV_{rms}}}  \\
					{\text{Static error}<0.1} &{\text{Phase Margin}>60 \mathrm{\enspace deg.}}\\
					
			\end{array}}
		\end{array}%
	$}
	\vspace*{0mm}
\end{equation}
\normalsize
In our experiment, the following transistors are required to operate in the saturation region: M1, M3, M4, M7, M9, M10, M12, M13, and [M15-M26]. The total number of design constraints becomes 29.

The statistical results for all the reference algorithms are shown in Table II. DNN-Opt shows high reliability and find a feasible solution in all its trials. However, other model-based methods, BO-wEI and GASPAD, fail to achieve similar behavior. DE can also find feasible results, but  DNN-Opt is 24x more efficient in the number of required simulations to find the first feasible result. It is also demonstrated in Table II that, on average, the final design proposed by DNN-Opt draws up to 43\% less power. The modeling time required by DNN-Opt is up to 50x smaller compared to other model-based methods. This results in 2.5--16x efficiency for total runtime.

\begin{figure}[t]
	\centering
	\includegraphics[scale=0.5]{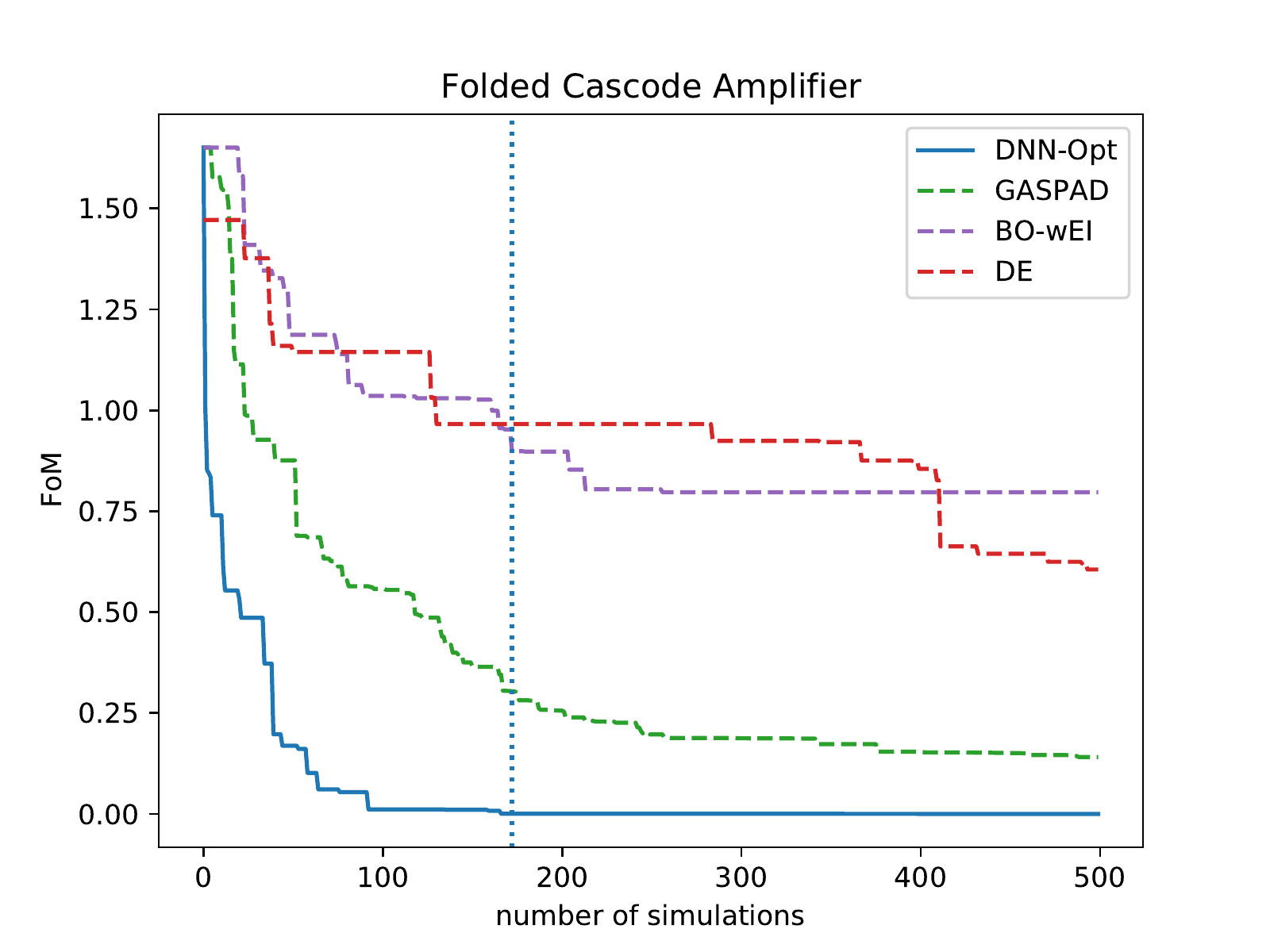}
	\vspace*{-3mm}
	\caption{The average FoM (lower is better) curve for 500 simulations}
	\label{fig:Folded_FoM}
	\vspace*{-2mm}
\end{figure}

\begin{table}
	\label{table:Folded_Results}
	\begin{centering}
		\caption{Statistics for different algorithms: Folded Cascode OTA}
	\end{centering}
	\label{table:foldedresults}
	\vspace*{-4mm}
	\begin{center}
		\resizebox{0.96\columnwidth}{!}{%
				\begin{tabular}{|l|cccc|}
					\hline
					Algorithm & DE & BO-wEI & GASPAD &\textbf{DNN-Opt} \\
					\hline
					success rate & 10/10 & 2/10 & 4/10 & \textbf{10/10}\\
					\hline
					\# of simulations & 3200 & $>$500 & $>$500 &  \textbf{132}\\
					\hline
					Min power (m$W$) & 0.75 & 0.91 & 0.72  & \textbf{0.62}\\
					\hline
					Max power (m$W$) & 1.53 & 1.62 & 1.75 & \textbf{0.77}\\
					\hline
					Mean power (m$W$) & 1.14 & 1.25 & 0.96  & \textbf{0.71}\\
					\hline
					Modeling time (h) &NA & 30 &6.5 & \textbf{0.6}\\
					\hline
					Simulation time (h) &54 &2.7 &2.7 &2.7 \\
					\hline
					Total runtime (h) &54 &32.7 &8.2 &\textbf{3.3} \\
					\hline
				\end{tabular}%
			}
		\end{center}
		\vspace{-7mm}
	\end{table}
	Figure \ref{fig:Folded_FoM} includes the FoM curve with iterations, where DNN-Opt shows strong convergence behavior and outperforms other methods. For our ten runs, DNN-Opt finds the feasible solution within $\text{205}$ iterations (marked with vertical dashed line) across all its ten trials. Although it is slow, GASPAD shows convergence to optimal FoM, but we observed that BO-wEI is often trapped in local optima. 
	
\begin{figure}[t!]
		\vspace*{0mm}
		\centering
		\vspace*{-3mm}
		\includegraphics[scale=0.45]{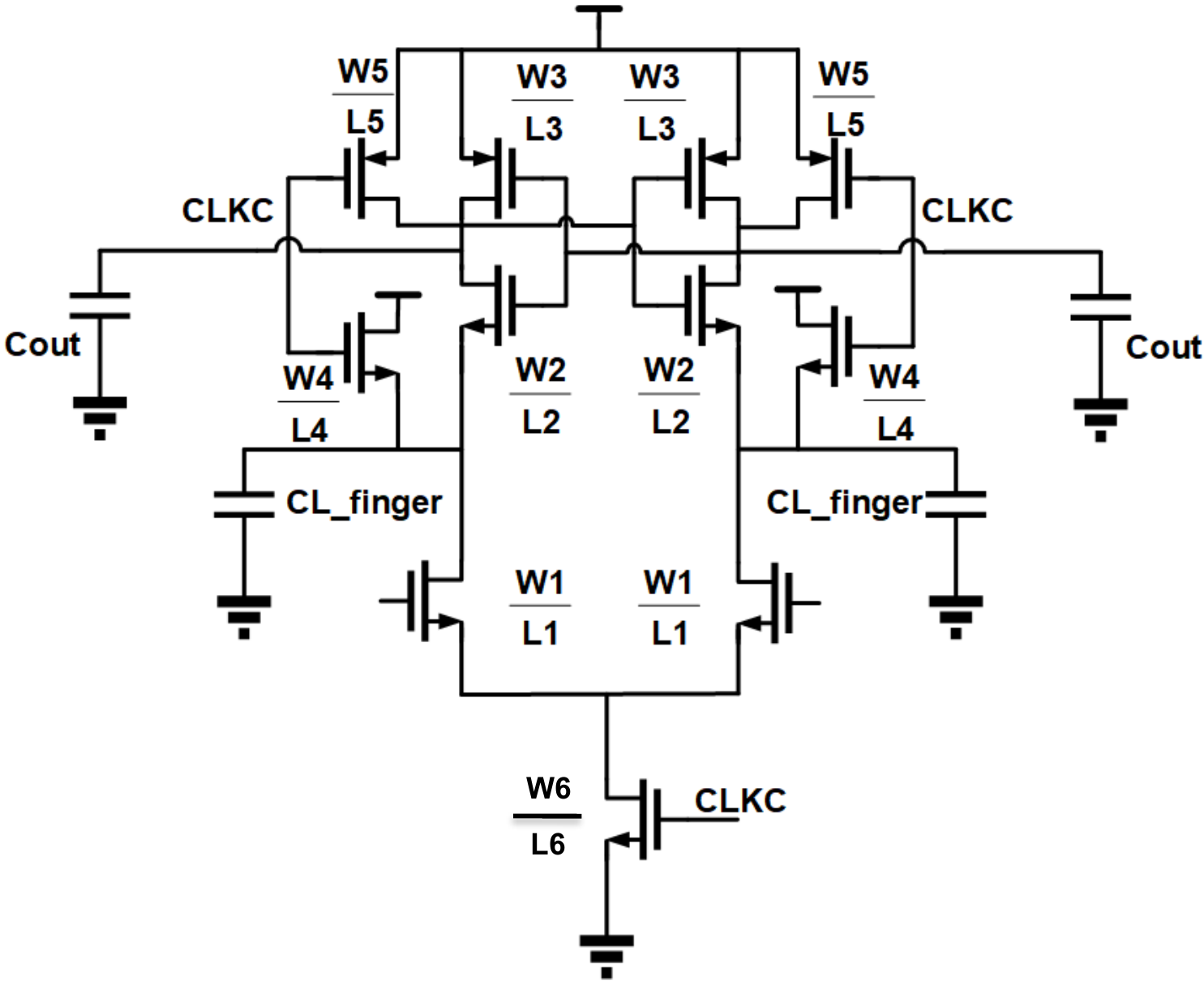}
		\vspace*{-2mm}
		\caption{Schematic of SA-Latch Comparator}
		\label{fig:SAschematic}
		\vspace*{-3mm}
	\end{figure}

	\textbf{Strong-Arm Latch Comparator}:
	The second test case is SA-Latch Comparator, which is shown in Figure \ref{fig:SAschematic}. It has 13 design variables, and their names and bounds are shown in table \ref{table:SAspace}.
	
	\vspace*{-3mm}
	\begin{table}[!h]
		\centering
		\caption{Design parameters and their ranges for SA-Latch Comparator}
		\label{table:SAspace}
		\vspace*{-3mm}
		\begin{center}
			\resizebox{0.9\columnwidth}{!}{%
				\begin{tabular}{l |l |l| l}
					\hline
					Parameter Name & Unit & LB & UB \\
					\hline
					L1-L2-L3-L4-L5-L6 \; \; \; \;  & $\mu\text{m}$& 0.18 & 10\\
					\hline
					W1-W2-W3-W4-W5-W6 \;\;\;\;\;\;\;\;\;\;\;& $\mu\text{m}$& 0.22 & 50\\
					\hline
					CL\_finger& integer & 10 & 300\\
					\hline
				\end{tabular}%
			}
		\end{center}
		\vspace*{-2mm}
		\vspace*{-2mm}
	\end{table}
	
	\begin{figure}[t!]
		\centering
		\includegraphics[scale=0.5]{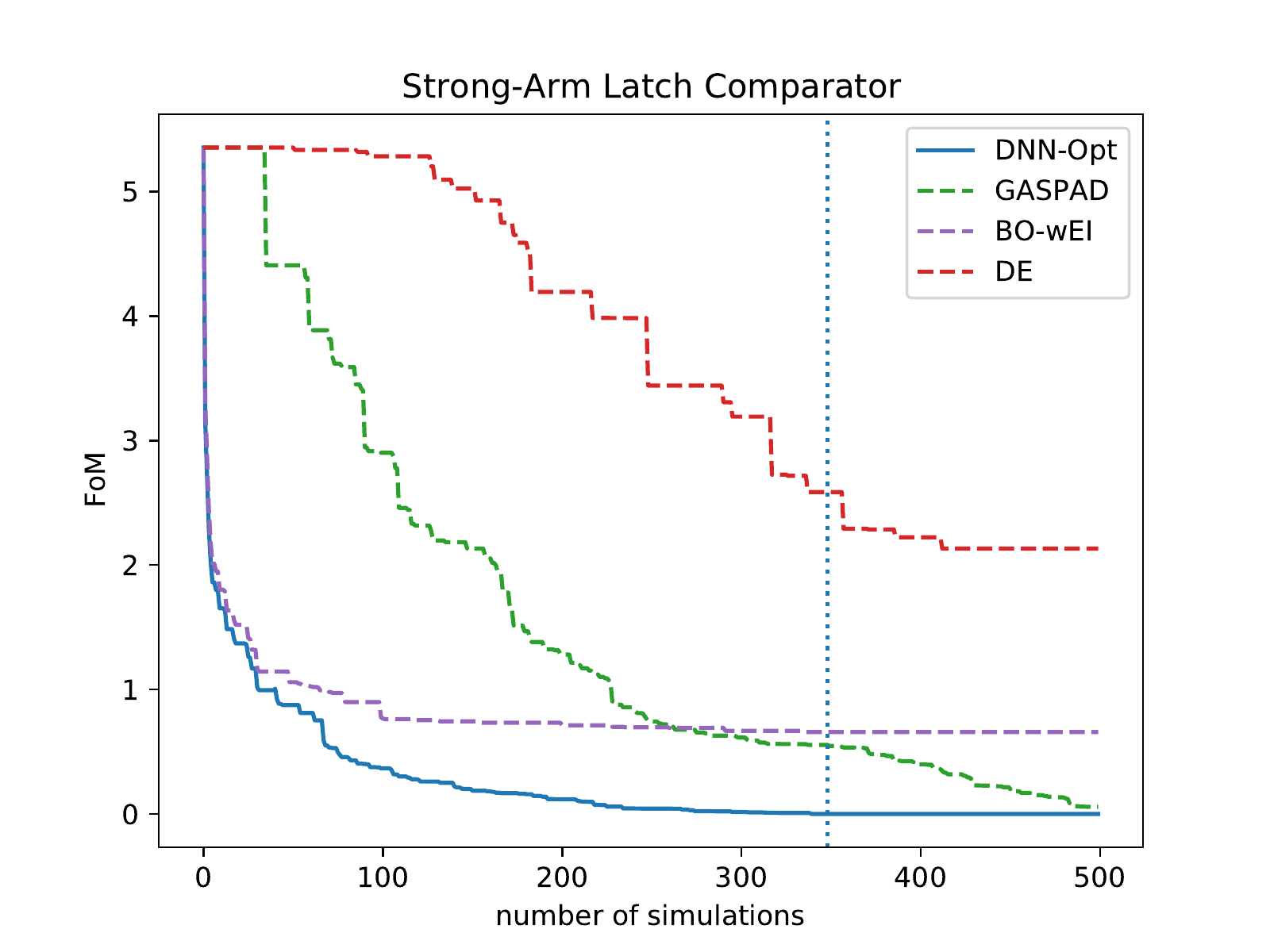}
		\vspace*{-3mm}
		\caption{The average FoM (lower is better) curve for 500 simulations}
		\label{fig:SA_FoM}
		\vspace*{0mm}
	\end{figure}

	The constrained optimization problem consists of 10 constraints in total:
	\small
	\begin{equation}
		\label{specifications:SA}
		\begin{array}{l}{\text { minimize } \text{Power}} \\
			{\text { s.t. } \text{\enspace Set Delay}<10 \mathrm{\enspace ns}} \\
			{\qquad \begin{array}{l}{\text{Reset Delay}<6.5 \mathrm{\enspace ns}} \\
					{\text{Area}< \mathrm{26} \mathrm{\enspace \mu m^2}} \\
					{\text{Input-referred Noise}< \mathrm{50} \mathrm{\enspace \mu Vrms}} \\
					{\text{Differential Reset Voltage}< \mathrm{1} \mathrm{\enspace \mu V}} \\
					{\text{Differential Set Voltage}> \mathrm{1.195} \mathrm{\enspace V}} \\
					{\text{Positive-Integration Node Reset Voltage}< \mathrm{60} \mathrm{\enspace \mu V}} \\
					{\text{Negative-Integration Node Reset Voltage}< \mathrm{60} \mathrm{\enspace \mu V}} \\
					{\text{Positive-Output Node Reset Voltage}< \mathrm{0.35} \mathrm{\enspace \mu V}} \\
					{\text{Negative-Output Node Reset Voltage}< \mathrm{0.35} \mathrm{\enspace \mu V.}} \\   
			\end{array}}
		\end{array}
		\vspace*{0mm}
	\end{equation}
	\normalsize 
	
	The statistical results for all the reference algorithms are shown in Table-\ref{table:SA_results}. Due to relatively tighter constraints for SA-Latch Comparator, methods typically needed a larger number of simulations to converge. DDN-Opt is the only method that finds a feasible solution in all trials, and our method shows more than 30x efficiency compared to DE. GASPAD shows relatively competitive results, but DNN-Opt finds a solution with 25\% better power consumption than successful runs of GASPAD. The runtime observations are similar to the folded cascode case.  
	
	FoM curves are shown in Figure \ref{fig:SA_FoM} for different methods. DNN-Opt finds a feasible solution within 348 simulations, which is much earlier than the others. BO-wEI shows a similar convergence trend for initial iterations then fails to model one of the constraints properly. Our observations showed that all the runs with the BO-wEI method were unable to meet input-referred noise, and some failed for set delay.

	\vspace{-2mm}
	\begin{table}[h!]
		\centering
		\caption{SA Latch Comparator Results}
		\vspace{-2mm}
		\label{table:SA_results}
		\resizebox{0.96\columnwidth}{!}{%
			\begin{tabular}{|c|c|c|c|c|}
				\hline
				Algorithm&DE&BO-wEI&GASPAD&DNN-Opt \\
				\hline
				success rate & 5/10  & 0/10 & 6/10 & \textbf{10/10}\\
				\hline
				\# of simulations & $>$10000 & $>$500 & $>$500 &  \textbf{330}\\
				\hline
				min ($\mu$ W) & 2.98 & NA & 3.05  & \textbf{2.50}\\
				\hline
				max ($\mu$ W) & 4.22 & NA & 3.75 & \textbf{2.75}\\
				\hline
				mean ($\mu$ W) & 3.57 & NA & 3.45  & \textbf{2.65}\\
				\hline
				Modeling time (h) &NA &17 & 3 & \textbf{0.3}\\
				\hline
				Simulation time (h) &72 &3.6 &3.6 &3.6 \\
				\hline
				Total runtime (h) & 72 &20.6 &6.6 &\textbf{3.9} \\
				\hline
			\end{tabular}%
		}
		\vspace{-5mm}
	\end{table}

\subsection{Experiments with Industrial Scale Circuits }

We tested DNN-Opt on four industrial circuits designed at a very advanced  technology node. These circuits were already in the process of manual sizing by expert analog designers and needed some fine-tuning. For these industrial circuits, we did not have access to other algorithms (DE, GASPAD, BO-wEI), and hence our baseline is with a commercial black-box optimizer based on Simulated Annealing. As will be demonstrated in this section, DNN-Opt performs well on large circuits and is not limited to small examples. Analog designers assisted in selecting permissible parameter ranges of the devices, considering layout impacts and process rules. For industrial cases, we identify critical devices based on Eq. \ref{eq:sens_equation} for the failing constraints ($f_i$'s of Eq. \ref{eq:prob_formulation}). Note, MLParest \cite{9218495} was used in the loop of DNN-Opt which helps analog designer estimate post-layout effects early in the design. 

\textbf{Inverter Chain}:
The first case is a simple inverter chain used mainly for tool development and flow testing. We used all the devices (8) in the four stage inverter chain. There were only two specs, delay and power. 

\textbf{Level Shifter}:
Sensitivity analysis identified ten critical devices impacting failing performances, and that led to a design space of $3.9\times 10^{15}$. There were 60 total specs like delay, rise, fall, power, current, etc.

\textbf{Low-Dropout (LDO) Regulator}:
We used sensitivity analysis to identify six critical devices leading to search space of $1.6\times 10^{13}$. The circuit had PSRR, Gain Margin, Phase Margin, DC Gain, GBW, etc., as part of nine constraints. The number of devices is high due to arrayed instances used by the analog engineer.

\textbf{Continuous-Time Linear Equalizer (CTLE)}:
Sensitivity analysis identified eight critical devices impacting failing performances. With design parameter and ranges identified by analog designers, we had a design space of $3.3\times 10^{25}$. There were a total of 14 constraints like DC Gain, offset, Nyquist Gain, Fpeak, Peaking Max, Power, etc. 

As illustrated in Table-V, DNN-Opt outperforms commercial optimizer available in the industry in terms of the number of simulations required to meet the constraints by 5x. We would like to emphasize that we can deal with fairly complex CTLE circuit by using 4x smaller number of costly SPICE simulations. Additionally, the optimal solution proposed by DNN-Opt consumed 8\% lesser power than simulated annealing. Our examples represent real use cases where designers already spend several days worth of human time in fixing constraints. Had we started with designs without any knowledge of human designers baked-in, we would have seen even greater returns in sample efficiency like Section \ref{sec:small_building_blocks}. 

\begin{table}
	\begin{centering}
		\caption{DNN-Opt Results on Industrial Circuits}
		\vspace{-3mm}
	\end{centering}
	\label{table:industrialcktresult}
	\begin{center}
		\resizebox{0.99\columnwidth}{!}{%
				\begin{tabular}{|l|c|c|c|c|}
					\hline
					Circuit & MOS & Nodes & Simulated Annealing (SA) &\textbf{DNN-Opt} \\
					\hline
					Inverter Chain & 8 & 7 & $>$1000 &\textbf{90}\\
					\hline
					Level Shifter & 1.2k & 3.9k & 1200 &\textbf{195}\\
					\hline
					LDO & 167k & 2.8k & 552 &\textbf{112}\\
					\hline
					CTLE & 173k & 63k & 587 &\textbf{150}\\
					\hline
				\end{tabular}%
			}
		\end{center}
		\vspace*{-3mm}
		\begin{center}
			\begin{tablenotes}
				\small
				\item Number of SPICE simulations shown in column SA and DNN-Opt for meeting constraints (lower is better).
			\end{tablenotes}
		\end{center}
		\vspace*{-2mm}
		\vspace{-4mm}
		
	\end{table}

\section{Conclusion
\label{sec:conclusions}}
In this work, we presented DNN-Opt, a novel sample efficient black-box optimization algorithm that combined the strengths of deep neural networks and reinforcement learning paradigm. We also give a recipe to extend our work for large circuits with thousands of devices. Our algorithm's effectiveness has been successfully demonstrated on various circuit building blocks and large industrial circuits leading to 5--30x sample efficiency, while being able to find feasible solution for all circuit sizing tasks and showing superior converge curves compared to other methods.
\section*{Acknowledgement}
This work is supported in part by NSF under Grant No. 1704758.
\bibliographystyle{IEEEtran}
\bibliography{bib/reference}

\end{document}